\documentclass[letterpaper]{article} 
\pdfoutput=1
\usepackage{aaai2027}
\usepackage[hyphens]{url}  
\usepackage{graphicx} 
\usepackage{natbib}  
\usepackage{caption} 
\usepackage{amsmath,amssymb}
\usepackage{booktabs}
\usepackage{multirow}
\usepackage{xspace}

\nocopyright

\newcommand{\Energy}{\mathcal{E}}
\newcommand{\Attribution}{\phi}
\newcommand{\Requests}{\mathcal{N}}
\newcommand{\Sset}{\mathcal{S}}
\newcommand{\Power}{P}
\newcommand{\Idle}{P_{\mathrm{idle}}}
\newcommand{\approach}{\textsc{JouleShare}\xspace}
\newcommand{\Calib}{\textsc{JCalib}\xspace}

\title{Request-Level Energy Attribution for Batched LLM Serving}
\author{
    Qi Luo\textsuperscript{\rm 1},
    Kunlin Li\textsuperscript{\rm 1},
    Ziwen Wang\textsuperscript{\rm 1},
    Dongsheng Wang\textsuperscript{\rm 2},
    Yun Chen\textsuperscript{\rm 1}\corresponding
}
\affiliations{
    \textsuperscript{\rm 1}The Hong Kong University of Science and Technology (Guangzhou)\\
    \textsuperscript{\rm 2}Tsinghua University\\
    \{qluo615, kli082, zwang578\}@connect.hkust-gz.edu.cn, wds@tsinghua.edu.cn, yunchen@hkust-gz.edu.cn
}

\begin{document}

\maketitle

\begin{abstract}
Batched LLM serving improves throughput but complicates energy accounting. GPU power telemetry is aggregate, whereas sustainability reporting, chargeback, and workload analysis often require request-level energy charges. Existing inference-energy benchmarks report model-, phase-, or token-level energy, and recent carbon-accounting work motivates Shapley fairness conceptually. Neither provides measured request-level ground truth, so how far the accounting rules used in practice deviate from a fair allocation has remained unknown. We present \approach, an attribution framework with two components. An offline harness establishes this ground truth by replaying request subsets under vLLM with a reproducible protocol, integrating GPU power telemetry, and computing exact Shapley energy for each request. A lightweight calibration model, \Calib, then learns to predict Shapley shares from cheap request features for use at serving time. Across 16 model/workload runs, token-proportional attribution differs from exact Shapley by 0.440 normalized L1 on average under static batching and by 0.458 under continuous batching, a gap that reproduces across three data-center GPUs. \Calib reduces this error to 0.116 under static batching and 0.177 under continuous batching, below even a standalone-measurement baseline that is unavailable online, while preserving exact batch-energy efficiency. Sampled Shapley extends the measured reference to larger group sizes, where the gap persists and a single offline calibration remains the most accurate deployable rule. The results show that token attribution is not a reliable proxy for marginal energy under batched execution, and that measured Shapley ground truth can calibrate low-cost request features toward fairer attribution.

\end{abstract}

\section{Introduction}

Production LLM deployments batch requests from many users to improve GPU utilization and amortize memory movement \citep{kwon2023vllm,zheng2024sglang,agrawal2023sarathi}. This batched execution raises a basic attribution question. When a batch consumes a certain number of joules, how many should be assigned to each request?

\begin{figure}[t]
\centering
\includegraphics[width=0.9\columnwidth]{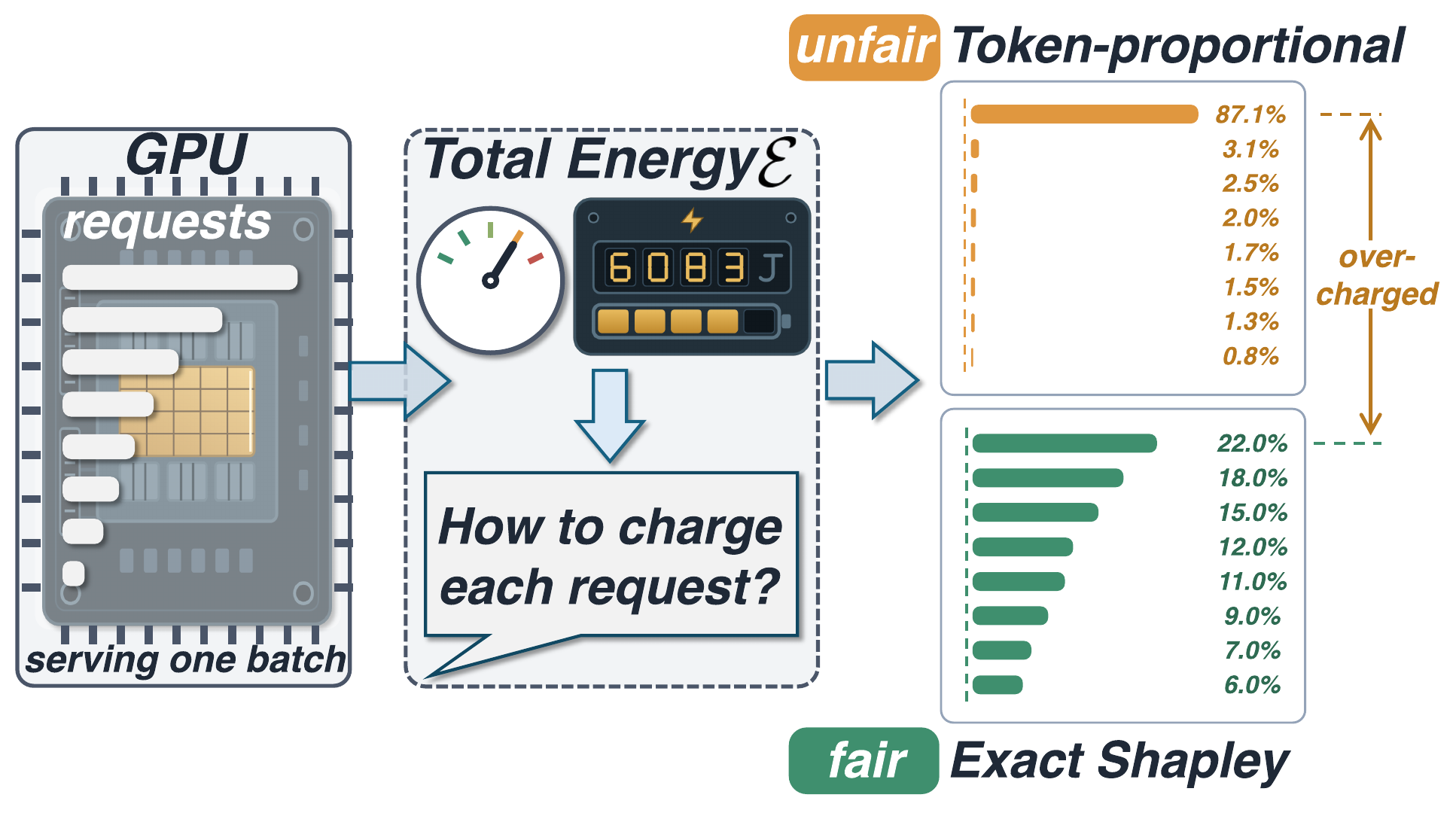}
\caption{Under batching, token attribution over-charges the long-prompt request relative to its exact Shapley share.}
\label{fig:teaser}
\end{figure}

Request-level attribution is needed for sustainability and cost reporting \citep{strubell2019energy,henderson2020systematic,schwartz2020greenai}, for identifying which workloads drive energy, and for assigning credit to serving optimizations. Token-proportional attribution splits batch energy in proportion to each request's token count \citep{chung2025mlenergy}, and standalone-proportional attribution splits it in proportion to each request's isolated energy, which requires serving every request alone and is impractical in production. Both rules are efficient in the sense that their charges sum to the observed batch energy, but neither accounts for how requests interact when served together. As Figure~\ref{fig:teaser} illustrates, a long-prompt request can receive a large token-proportional charge even when its marginal contribution under batching is much smaller.

We study request-level GPU energy attribution for LLM serving, taking active GPU energy from power telemetry as the measured signal. Given a set of co-served requests, the attribution vector should sum to the measured batch energy and assign each request a charge that reflects its average marginal contribution under batching. We use the Shapley value \citep{shapley1953value} as this fairness target because it is the canonical shared-cost allocation satisfying efficiency, symmetry, null-player, and linearity. Recent carbon-accounting work argues for Shapley values in multi-tenant LLM serving on conceptual and axiomatic grounds \citep{li2025llmcarbaccountant}. However, it does not measure request-level Shapley energy on a real serving stack, nor does it quantify how far simple rules such as token attribution depart from the Shapley allocation under batched execution.

The empirical challenge is that exact Shapley attribution requires the energy of every request subset, and nonlinear GPU utilization, heterogeneous prompts, variable decode lengths, and shared runtime overheads make this subset energy hard to model analytically \citep{zhong2024distserve,patel2023splitwise}. Our key observation is that this target can still be \emph{measured} for small request groups by replaying all subsets under a fixed serving protocol, both as static batches and under continuous batching with interleaved arrivals. The resulting Shapley ground truth both audits existing attribution rules and supervises a low-cost estimator.

We present \approach, an attribution framework that pairs an exact subset-replay harness with an offline-trained calibration model, \Calib, and we make the following contributions:

\begin{itemize}
    \item \textbf{Measured Shapley ground truth for LLM serving.} \approach replays every subset of a request group on vLLM with GPU power telemetry, under both static and continuous batching, and computes exact Shapley energy per request, so every energy value entering the Shapley computation is measured on a real serving stack rather than modeled.
    \item \textbf{An audit of token-proportional attribution.} Across 16 model/workload runs and 40,800 measured subset generations, token attribution differs from exact Shapley by 0.440 normalized L1 on average under static batching and by 0.458 under continuous batching, charging roughly a quarter of each batch's energy to the wrong requests. The gap is statistically significant and persists across group sizes, three data-center GPUs, and model scales from 1.5B to 14B.
    \item \textbf{A Shapley-aligned calibration model.} Trained on the measured attributions, \Calib recovers Shapley shares from lightweight request features to 0.116 normalized L1 under static batching and 0.177 under continuous batching, with a prediction overhead of about 0.003 ms per request. A single offline calibration transfers across GPUs and group sizes while remaining the most accurate deployable rule.
\end{itemize}

\section{Related Work}
\label{sec:related}

\paragraph{LLM inference energy measurement}
Recent work makes LLM inference energy observable, but usually at model, phase, or token granularity rather than as a request-level share of a shared batch \citep{niu2025tokenpowerbench}. ML.ENERGY reports per-request energy by dividing measured steady-state energy by token counts \citep{chung2025mlenergy}, the accounting that our token-proportional baseline audits. Other measurements split batch energy uniformly across prompts or predict per-prompt energy and carbon from request and hardware features \citep{caravaca2025prompts,fu2024llmco2}. The broader ML sustainability literature motivates energy and carbon reporting \citep{strubell2019energy,henderson2020systematic,schwartz2020greenai,patterson2021carbon,faiz2023llmcarbon}, but none of these works divides the measured energy of a shared batch among the requests inside it.

\paragraph{Carbon and energy attribution}
Attributing shared cost with the Shapley value has a long lineage, from airport landing fees to energy accounting in mobile systems and data centers \citep{littlechild1977aircraft,dong2014rethink,islam2016energy}. Sampling-based approximations reduce the number of coalition evaluations that exact Shapley needs \citep{castro2009polynomial}, but each coalition evaluation here is a physical measurement, so even approximate Shapley remains an offline procedure. Process-, VM-, and thread-level power meters attribute device energy by resource usage \citep{kansal2010joulemeter,he2023energat}, without modeling each tenant's marginal contribution. Recent systems apply related ideas to carbon and energy allocation in data centers and serverless functions \citep{han2025fairco2,rehman2024faasmeter}. The closest LLM work adopts Shapley as the fairness criterion for multi-tenant LLM serving, but supports the choice with conceptual and axiomatic examples rather than measured GPU energy on a real serving stack \citep{li2025llmcarbaccountant}. \approach instead measures the coalition energy function by exhaustively replaying request subsets under vLLM with GPU power telemetry, computes exact request-level Shapley energy under both static and continuous batching, and uses that measured reference to audit token-proportional attribution and train a lightweight estimator.

\paragraph{LLM serving systems}
Serving systems such as vLLM and SGLang improve throughput through PagedAttention, RadixAttention, and runtime scheduling \citep{kwon2023vllm,zheng2024sglang}, token-based fair scheduling regulates how requests share the batch \citep{sheng2024vtc}, and chunked prefill and phase disaggregation reshape the amount and timing of per-request work \citep{agrawal2023sarathi,patel2023splitwise,zhong2024distserve}. These mechanisms increase the interaction between requests, yet power telemetry still observes the aggregate batch, not the marginal energy of an individual request. \approach is complementary to cost-aware serving systems \citep{ecoserveorchestration2025}, providing the attribution substrate for identifying which requests drive energy and which serving optimizations should receive credit.

\section{Problem Formulation}
\label{sec:problem}

\subsection{Measured Batch Energy}

Let $\Requests=\{1,\ldots,n\}$ be a group of LLM inference requests. A request $i$ has prompt $x_i$ and prefill token count $p_i$. Its generated output $y_i$ and decode token count $d_i$ are produced by the serving system and, because generation is endogenous to batch composition, are not batch-invariant even in production serving. The player in the game below is therefore the prompt served under a fixed measurement protocol, with its output realized per coalition. Unless noted, $y_i$ and $d_i$ refer to the full-group serving. For a subset $\Sset \subseteq \Requests$, let $\Energy(\Sset)$ denote the active GPU energy consumed when serving exactly $\Sset$ under that protocol, with $\Energy(\emptyset)=0$.

In our implementation, $\Energy(\Sset)$ is computed from GPU power samples:
\begin{equation}
    \Energy(\Sset) =
    \int_{t_0}^{t_1} \max(\Power_\Sset(t) - \Idle, 0) \, dt,
\end{equation}
where $\Power_\Sset(t)$ is the GPU power sampled while serving $\Sset$, $\Idle$ is the measured idle GPU power, and $[t_0,t_1]$ is the measurement window. Subtracting $\Idle$ isolates the \emph{dynamic} energy that serving adds on top of the device's always-on static draw, which is independent of which requests run and should not be charged to any of them. The $\max(\cdot,0)$ clamps occasional non-physical samples that dip below $\Idle$ from sensor noise during the window's short idle padding intervals. We treat GPU active energy as the primary signal. Operational carbon follows by multiplying energy by a carbon-intensity factor, so an energy attribution induces the same carbon shares, and embodied carbon is beyond our scope.

We measure $\Energy(\Sset)$ in two serving regimes: static batches, where $\Sset$ is served as a single batch, and continuous batching (CB), where the requests of $\Sset$ are injected into an iteration-level scheduler at fixed, interleaved arrival times. Both are counterfactual replays of the same coalition under a reproducible protocol, so the Shapley construction below applies unchanged in either regime. \S\ref{sec:experiments} compares the two.

\subsection{Attribution Target}

An attribution method returns a vector $\Attribution(\Requests) \in \mathbb{R}^n$, where $\Attribution_i(\Requests)$ is the energy charged to request $i$. We call a method \emph{efficient} if
\begin{equation}
    \sum_{i \in \Requests} \Attribution_i(\Requests) = \Energy(\Requests).
\end{equation}
Efficiency is necessary, guaranteeing that no energy disappears and no extra energy is invented. However, it alone does not determine how energy should be distributed. Batched serving makes the split hard, because energy is nonlinear in token counts. Batching changes GPU utilization, amortizes shared overhead, and makes each request's cost depend on its companions, so a request's true cost is its marginal contribution. We use the Shapley value as the fairness target:
\begin{equation}
\begin{aligned}
    \Attribution_i^{\mathrm{Shapley}}(\Requests)
    ={}& \sum_{\Sset \subseteq \Requests \setminus \{i\}}
    \frac{|\Sset|!(n-|\Sset|-1)!}{n!} \\
    &{}\cdot
    \left[
        \Energy(\Sset \cup \{i\}) - \Energy(\Sset)
    \right].
\end{aligned}
\label{eq:shapley}
\end{equation}
This value is the average marginal energy contribution of request $i$ over all hypothetical orders in which requests could join the group, distinct from the physical arrival order. Exact Shapley attribution serves as our measured reference. Any request-level Shapley ground truth is necessarily counterfactual, since $\Energy(\Sset)$ for $\Sset\neq\Requests$ is never observed in production. A reproducible replay protocol makes it measurable. We compute it offline and use it to audit existing rules and train \Calib.

\subsection{Attribution Rules}

We compare exact Shapley against three simpler rules.

\paragraph{Token-proportional attribution (Token)}
Energy is split in proportion to total tokens:
\begin{equation}
    \Attribution_i^{\mathrm{Token}}(\Requests)
    =
    \Energy(\Requests)
    \frac{p_i + d_i}{\sum_{j \in \Requests}(p_j+d_j)}.
\end{equation}
This rule extends ML.ENERGY's token-normalized accounting to shared batches by allocating energy in proportion to each request's prompt and decode tokens \citep{chung2025mlenergy}.

\paragraph{Standalone-proportional attribution (Solo)}
The batch energy is split in proportion to each request's isolated energy:
\begin{equation}
    \Attribution_i^{\mathrm{Solo}}(\Requests)
    =
    \Energy(\Requests)
    \frac{\Energy(\{i\})}{\sum_{j \in \Requests}\Energy(\{j\})}.
\end{equation}
This rule is more accurate than token attribution but needs an isolated run per request, so it serves as an offline reference.

\paragraph{\Calib}
Our calibration model predicts Shapley shares from cheap per-request features. \S\ref{sec:approach} details its design.

\begin{figure*}[t]
\centering
\includegraphics[width=0.9\textwidth]{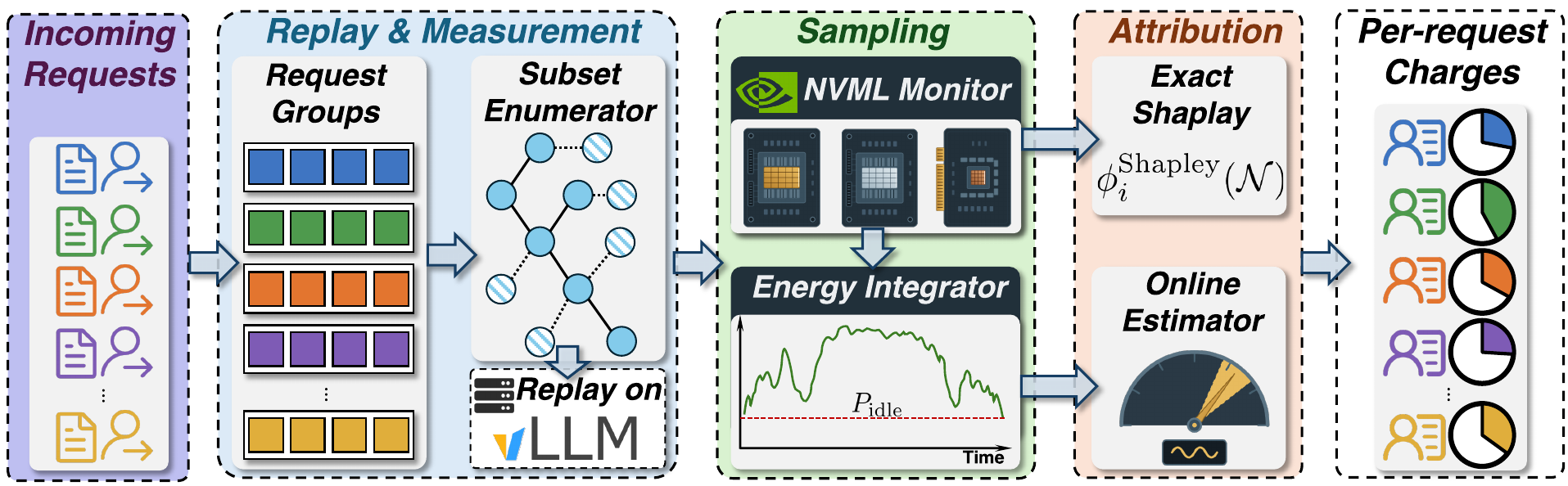}
\caption{\approach architecture. The offline harness replays request subsets under vLLM with GPU power telemetry to compute exact Shapley energy, which supervises \Calib for attribution at serving time.}
\label{fig:overview}
\end{figure*}

\section{\approach}
\label{sec:approach}

\approach separates the expensive part of attribution from the part that must run online. The offline harness measures the coalition energy function directly on the serving stack and computes exact Shapley attributions from it. \Calib then trains on these measured attributions and, at serving time, produces efficient request-level charges from lightweight features. Figure~\ref{fig:overview} summarizes this pipeline.

\subsection{Exact Shapley Measurement Harness}

For a group of $n$ requests, exact Shapley computation requires $\Energy(\Sset)$ for every non-empty subset $\Sset \subseteq \Requests$. This stage is exponential in $n$ and is intended for offline ground-truth construction. \approach automates the exhaustive measurement with four components.

\paragraph{Workload construction}
The harness builds heterogeneous request groups from four workload families: \textbf{Chat}, conversational prompts drawn from the public ShareGPT corpus \citep{sharegpt2023}; \textbf{LongBench}, long-context prompts with heterogeneous prompt and output lengths \citep{bai2023longbench}; \textbf{Reasoning}, GSM8K and MATH step-by-step prompts \citep{cobbe2021gsm8k,hendrycks2021math}; and \textbf{Synthetic}, controlled short/medium/long prompt and output-instruction combinations. Each group spans the full range of prompt lengths in its workload.

\paragraph{vLLM replay}
Each subset is replayed through vLLM with greedy decoding. Greedy decoding removes sampling randomness, while batch composition still affects runtime behavior and even generated lengths, effects that the measured coalition energy is meant to capture and that \S\ref{sec:experiments} quantifies. The same replay runs under static batching and under continuous batching with interleaved arrivals at a fixed 0.5~s gap, giving measured coalition energies in both regimes.

\paragraph{Power measurement}
For each subset, \approach starts an NVML power monitor that samples GPU power every 100~ms, waits for a short padding interval, and runs generation. It then subtracts the measured idle power from each sample and integrates active energy over the measurement window.

\paragraph{Exact attribution}
After subset runs complete, \approach averages any repeated measurements of each subset and applies Eq.~\ref{eq:shapley}. The output is a per-request attribution table with prefill and decode token counts, singleton energy, and the exact Shapley, Token, and Solo charges. These measured attributions are the reference used in the experiments.

\subsection{Shapley-Aligned Calibration Model}

Subset replay is unavailable at serving time, so \Calib predicts the measured Shapley allocation from features the serving system already has. For request $i$ in a group, the prediction target is its Shapley energy \emph{share} $s_i = \Attribution_i^{\mathrm{Shapley}}(\Requests)/\Energy(\Requests)$, which sums to one over the group. Table~\ref{tab:calib-features} lists the features, derived from per-request token counts and group aggregates. We keep prefill and decode as separate channels rather than collapsing them into a single total, because the two phases contribute asymmetrically to energy. This separation is the main correction to token-proportional attribution, and \S\ref{subsec:calib} shows that it also supports cross-model and cross-workload transfer.

\begin{table}[t]
\centering
\setlength{\tabcolsep}{5pt}
\footnotesize
\begin{tabular}{ll}
\toprule
\textbf{Feature group} & \textbf{Per-request features} \\
\midrule
Raw counts & prefill $p_i$, decode $d_i$, total $p_i{+}d_i$ \\
Log counts & $\log(1{+}p_i)$, $\log(1{+}d_i)$, $\log(1{+}p_i{+}d_i)$ \\
Within-group share & $p_i/\!\sum p$, $d_i/\!\sum d$, $(p_i{+}d_i)/\!\sum(p{+}d)$ \\
Ratio to group mean & $p_i/\bar p$, $d_i/\bar d$, $(p_i{+}d_i)/\overline{(p{+}d)}$ \\
\bottomrule
\end{tabular}
\caption{Features of the \Calib estimator, from per-request token counts $p_i$, $d_i$ and group aggregates.}
\label{tab:calib-features}
\end{table}

We standardize the features, stack the per-request feature vectors $x_i$ (with an intercept) as the rows of $X$ and the target shares as $s$, and fit ridge regression \citep{hoerl1970ridge} with $\lambda{=}1$ in closed form,
\begin{equation}
    \label{eq:ridge}
    \beta = (X^\top X + \lambda \tilde{I})^{-1} X^\top s,
\end{equation}
where $\tilde{I}$ zeroes the intercept entry of the identity so that only non-intercept weights are penalized. At serving time we score each request as $\hat{s}_i = \beta^\top x_i$, clip the scores to be non-negative, and renormalize them so that $\sum_i \hat{\Attribution}_i = \Energy(\Requests)$ exactly. This renormalization makes \Calib efficient by construction, while the ridge model learns a linear mapping over nonlinear token-count features. The model is a single weight vector, so prediction costs about $0.003$~ms per request and needs no GPU.

\Calib uses no singleton energy feature, since isolated measurement is unavailable at serving time. \S\ref{subsec:calib} reports its accuracy and generalization.

\paragraph{Attribution windows at serving time}
In live serving, requests arrive and depart continuously, so the group $\Requests$ and the batch energy $\Energy(\Requests)$ that renormalization targets are defined per accounting window. The serving system integrates GPU power over fixed wall-clock windows or scheduler epochs, and the requests active in a window form its group. \Calib scores each from the token-count features accumulated within the window, and renormalizing those scores divides the window's measured energy. A request spanning several windows is charged in each, so its total charge is the sum of its per-window shares and efficiency holds window by window. The 0.003 ms figure above is the model-scoring cost within a window. Power integration and token counting reuse telemetry the serving stack already collects. Our evaluation replays fixed groups, each corresponding to one window. Validating windowed attribution on organic arrival traces remains future work (\S\ref{sec:experiments}, Limitations).

\section{Experiments}
\label{sec:experiments}

We first quantify how far the Token and Solo rules deviate from exact Shapley attribution, in both the static and continuous-batching regimes (\S\ref{subsec:main}, with a worst-case worked example in Appendix~A.3). We then evaluate how accurately \Calib recovers Shapley attributions from cheap request features (\S\ref{subsec:calib}). Finally, we examine whether the token--Shapley gap persists under measurement noise, larger group sizes, different GPUs, and smaller models (\S\ref{subsec:groupsize} and \S\ref{subsec:robustness}).

\begin{table*}[t]
\centering
\small
\setlength{\tabcolsep}{6pt}
\begin{tabular}{llrrrrrrr}
\toprule
\multirow{2}{*}{\textbf{Model}} & \multirow{2}{*}{\textbf{Workload}} & \multirow{2}{*}{\textbf{Prefill}} & \multicolumn{3}{c}{\textbf{Static Batching}} & \multicolumn{3}{c}{\textbf{Continuous Batching}} \\
\cmidrule(lr){4-6} \cmidrule(lr){7-9}
 & & & \textbf{Decode} & \textbf{Token L1} & \textbf{Solo L1} & \textbf{Decode} & \textbf{Token L1} & \textbf{Solo L1} \\
\midrule
\multirow{4}{*}{Qwen2.5-7B} & Chat & 72 & 421 & 0.108 & 0.040 & 419 & 0.149 & 0.181 \\
 & LongBench & 921 & 253 & 0.546 & 0.150 & 253 & 0.605 & 0.175 \\
 & Reasoning & 103 & 273 & 0.268 & 0.239 & 275 & 0.285 & 0.283 \\
 & Synthetic & 1722 & 198 & 0.917 & 0.135 & 199 & 0.892 & 0.162 \\
\midrule
\multirow{4}{*}{Qwen2.5-14B} & Chat & 72 & 399 & 0.101 & 0.055 & 402 & 0.123 & 0.149 \\
 & LongBench & 921 & 243 & 0.607 & 0.214 & 246 & 0.622 & 0.161 \\
 & Reasoning & 103 & 299 & 0.264 & 0.215 & 301 & 0.261 & 0.251 \\
 & Synthetic & 1722 & 189 & 0.925 & 0.125 & 192 & 0.914 & 0.132 \\
\midrule
\multirow{4}{*}{Llama-3.1-8B} & Chat & 79 & 435 & 0.091 & 0.025 & 438 & 0.138 & 0.166 \\
 & LongBench & 1005 & 267 & 0.555 & 0.143 & 276 & 0.558 & 0.182 \\
 & Reasoning & 109 & 239 & 0.286 & 0.254 & 244 & 0.352 & 0.349 \\
 & Synthetic & 1053 & 233 & 0.815 & 0.156 & 235 & 0.797 & 0.211 \\
\midrule
\multirow{4}{*}{Mistral-7B} & Chat & 54 & 395 & 0.162 & 0.105 & 391 & 0.223 & 0.236 \\
 & LongBench & 1341 & 228 & 0.631 & 0.297 & 231 & 0.640 & 0.220 \\
 & Reasoning & 84 & 265 & 0.282 & 0.291 & 255 & 0.252 & 0.282 \\
 & Synthetic & 1702 & 316 & 0.476 & 0.163 & 324 & 0.510 & 0.167 \\
\midrule
Mean & All & 691 & 291 & 0.440 & 0.163 & 293 & 0.458 & 0.207 \\
\bottomrule
\end{tabular}
\caption{Normalized L1 against exact Shapley for the Token and Solo rules under static and continuous batching, averaged over five groups per model/workload row. Prefill and Decode are mean prompt and generated token counts per request.}
\label{tab:main-errors}
\end{table*}

\subsection{Experimental Setup}

\paragraph{Hardware and software}
Unless noted otherwise, the experiments ran on a single NVIDIA A800 80 GB GPU. \S\ref{subsec:robustness} extends the measurements to A40 and H100 GPUs. The serving engine was vLLM \citep{kwon2023vllm}. GPU power was sampled every 100 ms during each subset run with NVIDIA NVML. On each GPU we measured idle power before each campaign and subtracted its mean from every trace.

\paragraph{Models and workloads}
Our experiments cover four model variants: Qwen2.5-7B-Instruct, Qwen2.5-14B-Instruct \citep{qwen2024qwen25}, Llama-3.1-8B-Instruct \citep{grattafiori2024llama3}, and Mistral-7B-Instruct-v0.3 \citep{jiang2023mistral}, each evaluated on the four workload families from \S\ref{sec:approach}. \S\ref{subsec:robustness} adds Qwen2.5-1.5B-Instruct as a smaller-scale check.

\paragraph{Exact Shapley protocol}
Each exact group has $n=8$ requests, so a complete group requires $2^8-1=255$ subset generations. Each model/workload run uses 5 groups, giving 1,275 measured subset generations per run. The campaign spans 16 model/workload runs and 80 complete groups, each measured under both static and continuous batching with an identical protocol, for 40,800 subset generations in total. Decoding is greedy with \texttt{max\_tokens=512}.

\paragraph{Characteristic function and determinism}
We define $\Energy(\Sset)$ operationally as the measured active energy of serving $\Sset$ under this protocol. Even under greedy decoding, batched execution is not bitwise identical to isolated execution, so a request's generated length varies with batch composition. Across the subsets containing a request, its output length is constant for only 33\% of occurrences and otherwise varies by a median of 36 and at most 305 tokens (median 12 overall). This drift perturbs the ground truth only at second order. Repricing every subset at its fitted per-decode-token energy, as if all coalitions served the full-group lengths, moves the exact Shapley shares by a mean of 0.036 normalized L1 under static batching and 0.026 under CB, an order of magnitude below the Token gap. The same effect makes mean decode lengths differ slightly between the two regimes, which are measured separately (Table~\ref{tab:main-errors}). We compute Token attribution from each request's tokens as served in the full group.

\paragraph{Baselines and metrics}
The Token and Solo rules from \S\ref{sec:problem} serve as baselines. Our main metric is normalized L1 against exact Shapley,
\begin{equation}
    \label{eq:nl1}
    \frac{\sum_i \left| \hat{\Attribution}_i - \Attribution_i^{\mathrm{Shapley}}(\Requests) \right|}{\Energy(\Requests)}.
\end{equation}
Absolute energies in Joules accompany the analysis in \S\ref{subsec:main} and Appendix~A.1.

\subsection{The Token--Shapley Gap}
\label{subsec:main}

\paragraph{Batching savings}
Serving the eight requests together instead of one at a time uses a mean of 1989 J versus 8532 J under static batching, a 77\% saving, and 2545 J versus 8286 J under continuous batching, a 69\% saving (full energy breakdown in Appendix~A.1). The saved energy is shared, so the central question is how to divide the measured batch total.

\paragraph{Token misattribution}
Table~\ref{tab:main-errors} reports the attribution errors. Token attribution is consistently inaccurate when heterogeneous requests share a batch, with a mean normalized L1 of 0.440 under static batching. The error is smallest for chat, where token and runtime behavior are relatively homogeneous, and much larger for the long-context and synthetic groups, so the aggregate gap scales with the heterogeneity of the served mix. Replaying the same requests under continuous batching raises the mean to 0.458, so the gap is not an artifact of static replay and, if anything, grows under a live scheduler.

\paragraph{Per-request extremes}
The most mischarged request in a group is off by 0.207 of batch energy per static run on average, reaching 0.431 on Qwen2.5-14B LongBench, where a single 8254-prefill request is charged 2527.6 J by token attribution but contributes only 974.6 J by exact Shapley. Under continuous batching its Shapley value is nearly unchanged at 980.1 J, so the overcharge follows from shared batched execution, not the replay protocol. Appendix~A.3 gives the full group in both regimes.

\paragraph{Accuracy versus deployability}
Standalone-proportional attribution is a stronger offline baseline because it uses singleton measurements. Its mean L1 is 0.163 under static batching and 0.207 under continuous batching, well below token attribution in both regimes. However, singleton measurements require serving each request in isolation, which is unavailable in ordinary online serving. Token attribution is therefore deployable but inaccurate, while standalone attribution is accurate but not deployable. The calibration model below aims to be both.

\paragraph{Statistical significance}
To confirm that these gaps exceed sampling variation across the 80 groups per regime, we compute bootstrap 95\% confidence intervals and a paired permutation test on per-group L1. The Token and Solo intervals are well separated in both regimes, and the paired Token$-$Solo difference is significant at $p<10^{-4}$, 0.277 under static batching and 0.251 under CB. Appendix~A.2 lists the full intervals.

\subsection{Calibration Model Accuracy and Overhead}
\label{subsec:calib}

\paragraph{Accuracy}
\Calib is trained and evaluated on disjoint request groups within each model/workload run (protocol in Appendix~A.4). As shown in Figure~\ref{fig:online}, its mean L1 is 0.116 under static batching and 0.177 under continuous batching, versus 0.440 and 0.458 for token attribution, at about 0.003 ms per request. It also improves on Solo, whose mean L1 is 0.163 and 0.207, despite using no singleton measurements. Gains are largest on LongBench and synthetic, where token counts are most misleading. The improvement over both baselines is statistically significant in both regimes (Appendix~A.2).

\begin{figure}[t]
\centering
\includegraphics[width=\columnwidth]{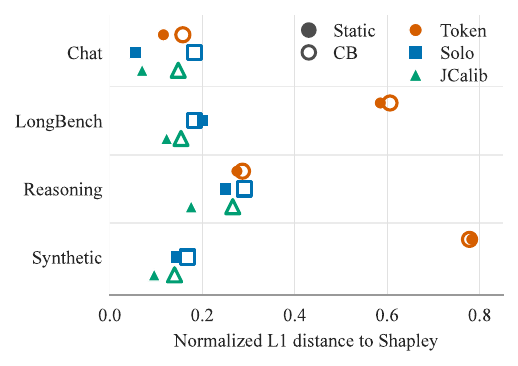}
\caption{Normalized L1 against exact Shapley for the three attribution rules, per workload, under static and continuous batching.}
\label{fig:online}
\end{figure}

\paragraph{Generalization}
The learned correction is not specific to one model/workload run. \Calib reaches 0.127 mean L1 on unseen models and 0.185 on unseen workloads, far below token attribution. The transferable signal is the separated prefill and decode channels. Collapsing them into a single total inflates the unseen-model error to 0.395. The same pattern holds under continuous batching, and a calibrator fit only on static runs reaches 0.216 on the CB measurements. Appendix~A.4 gives the full per-split numbers for both regimes.

\paragraph{Importance of Shapley supervision}
A per-phase cost model could avoid subset replay entirely, fitting $\Energy(\Requests) \approx a\sum_i p_i + b\sum_i d_i$ from full-batch energies alone and attributing in proportion to $a\,p_i + b\,d_i$. This rule reaches only 0.367 L1 under static batching and 0.335 under CB, barely better than token attribution. With the same two features, switching the supervision to measured Shapley shares improves L1 to 0.129 and 0.199. The gain comes from the supervision signal, not the feature set (Appendix~A.5).

\subsection{Sensitivity to Group Size}
\label{subsec:groupsize}

\begin{figure}[t]
\centering
\includegraphics[width=\columnwidth]{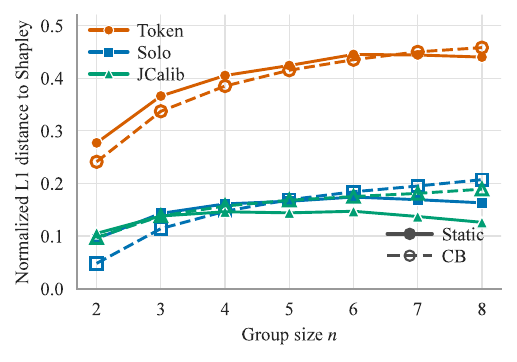}
\caption{Normalized L1 against exact Shapley versus group size $n$ for the three attribution rules under static and continuous batching, induced from the $n{=}8$ measurements.}
\label{fig:group-size}
\end{figure}

\paragraph{Induced sub-batches}
The measured subsets already contain every sub-batch, so we induce exact Shapley for every group size $n\in\{2,\ldots,8\}$ at no additional cost, averaging over all $\binom{8}{n}$ sub-batches of each 8-request group. Figure~\ref{fig:group-size} tracks the three rules as $n$ grows, under both static and continuous batching. Token L1 climbs monotonically, from 0.277 at $n=2$ through 0.445 at $n=6$, then plateaus near 0.44, so token attribution becomes a worse proxy as more heterogeneous requests share a batch. The continuous-batching curve climbs the same way and keeps rising to 0.458 at $n=8$. \Calib remains comparatively low across these induced sub-batch sizes (around 0.10 to 0.19 in both regimes), matching the impractical standalone baseline with only cheap features.

\paragraph{\boldmath Scaling to $n{=}16$ with sampled Shapley}
Exact enumeration is infeasible beyond small groups, so we extend the measured reference to 16-request groups with permutation-sampled Shapley \citep{castro2009polynomial}, running the campaign on the H100 for its throughput. For two new groups per workload, drawn from the same pools, we measure every prefix coalition along $m{=}200$--$400$ antithetic permutations per group and regime (83,456 replays in total), with larger $m$ where per-permutation variance is highest, and estimate each request's value as its mean marginal energy. The exact $n{=}8$ groups measured on the same GPU provide a same-hardware anchor, reported alongside the sampled results in Table~\ref{tab:sampled-n16}.

Because this reference is estimated from sampled permutations rather than a full enumeration, every error measured against it includes sampling noise. We quantify the noise by resampling the $m$ permutations with replacement and recomputing the estimate. The \emph{sampling floor} is the mean normalized L1 between the resampled and the full estimates, the error that an oracle returning the exact Shapley values would still appear to make. Measured errors therefore tend to overstate true ones by roughly the floor, and differences smaller than the floor are not resolved.

\begin{table}[t]
\centering
\setlength{\tabcolsep}{6pt}
\small
\begin{tabular}{llrrr}
\toprule
\textbf{Regime} & \textbf{Reference} & \textbf{Token} & \textbf{Solo} & \textbf{\Calib} \\
\midrule
\multirow{2}{*}{Static} & Exact, $n{=}8$ & 0.501 & 0.148 & \textbf{0.140} \\
 & Sampled, $n{=}16$ & 0.475 & 0.218 & \textbf{0.197} \\
\midrule
\multirow{2}{*}{CB} & Exact, $n{=}8$ & 0.508 & 0.226 & \textbf{0.196} \\
 & Sampled, $n{=}16$ & 0.512 & 0.325 & \textbf{0.283} \\
\bottomrule
\end{tabular}
\caption{Mean normalized L1 on the H100 (Qwen2.5-7B) against exact ($n{=}8$) and sampled ($n{=}16$) Shapley references.}
\label{tab:sampled-n16}
\end{table}

Table~\ref{tab:sampled-n16} reports the results, with the per-workload breakdown in Appendix~A.7. Mean Token L1 moves from 0.501 at $n{=}8$ to 0.475 at $n{=}16$ under static batching and from 0.508 to 0.512 under CB, so doubling the group size leaves the gap at its plateau, now by direct measurement rather than induction. The CB replays are saturated, sustaining at least 8.9 of 16 requests in flight on average. Solo degrades to 0.218 and 0.325 as companion interactions strengthen. \Calib is calibrated on A800 $n{=}8$ groups of the matching regime and applied unchanged, so it has seen neither the target GPU nor the target group size. It nevertheless reaches a measured L1 of 0.197 (sampling floor 0.079) under static batching and 0.283 (floor 0.044) under CB, the lowest mean deployable error in both regimes, so a single offline calibration transfers across both hardware and scale. Under static batching its apparent growth from $n{=}8$ is smaller than the sampling floor, so scale alone costs little. The residual CB growth is a regime effect rather than a scale effect. Fitted on static replays instead, the same calibrator degrades to 0.369 on the CB groups, and adding $n{=}16$ static groups to its training brings no gain.

\subsection{Measurement Repeatability and Robustness}
\label{subsec:robustness}

\paragraph{Repeatability}
To confirm the gap exceeds measurement noise, we re-measured eight Qwen2.5-7B groups, two per workload family, three times per subset. Subset energy is stable (per-subset CV $\sim$1\%) and decoding is repeatable, with output identical across repeats for 97.8\% of subsets, so the characteristic function is well defined. Token L1 varies by only $\pm0.007$ across repeats, roughly $69\times$ below the gap.

\paragraph{Cross-hardware}
Measuring the same Qwen2.5-7B groups on two further GPUs, an NVIDIA A40 48 GB and an NVIDIA H100 SXM5 80 GB, reproduces the gap on all four workloads. Mean Token L1 is 0.467, 0.475, and 0.501 on the A800, A40, and H100 under static batching, and 0.491, 0.458, and 0.508 under continuous batching. The misattribution is therefore not specific to the A800 or to one GPU generation (per-workload results in Appendix~A.6).

\paragraph{Model scale}
The gap also persists on Qwen2.5-1.5B, where mean Token L1 stays at 0.454 in both regimes and \Calib remains the most accurate deployable rule (per-workload results in Appendix~A.8). Token misattribution is thus robust across measurement repeats, GPUs, group sizes up to $n{=}16$, model scales (1.5B--14B), and batching regimes.

\subsection{Limitations}

\textbf{Energy coverage.} We report GPU rather than full-system power. The GPU dominates single-node power at 60--75\% \citep{Dodge2022CarbonCloud,patel2024characterizing,Yu2025SingleNodePower,AquinoBritez2025EnergyIndex}. \textbf{Measurement scope.} Both regimes define the characteristic function through counterfactual replay rather than organic production traffic, and validation on live arrival traces remains future work. Stage-level prefill and decode energy is not separated, as distinct from \Calib's token-count features, because the two stages interleave within an iteration. \textbf{Scale.} Exact measurement stops at $n{=}8$, sampled Shapley extends the reference to $n{=}16$ with a quantified sampling floor (\S\ref{subsec:groupsize}), and models stay at or below 14B on a single GPU. The measured plateau through $n{=}16$ shows that token misattribution does not vanish in larger groups, while the higher CB error leaves further scale validation necessary for the estimator.

\section{Conclusion}

Batching creates shared energy whose distribution across requests is not captured by token counts. \approach measures this attribution target directly by replaying request subsets under vLLM and integrating active GPU energy to compute exact Shapley attributions. Across 16 model/workload runs, token-proportional attribution differs from exact Shapley by 0.440 normalized L1 under static batching and 0.458 under continuous batching, a gap that reproduces across three data-center GPUs and model scales. \Calib, trained on the measured attributions with channel-separated features, recovers Shapley shares to 0.116 and 0.177 in the two regimes at about 0.003 ms per request. Permutation sampling extends the measured reference to larger group sizes, where the token gap does not shrink and the same offline calibration continues to give the most accurate deployable attribution. Measured coalition energy therefore provides a practical reference for auditing and calibrating request-level energy attribution in batched LLM serving.

\bibliography{references}

\appendix
\section{Additional Experimental Details}

\subsection{Full Energy Breakdown}
\label{app:main-full}

\begin{table*}[t]
\centering
\small
\setlength{\tabcolsep}{5pt}
\begin{tabular}{clrrrrrrrrr}
\toprule
\multirow{2}{*}{\textbf{Model}} & \multirow{2}{*}{\textbf{Workload}} & \multicolumn{2}{c}{\textbf{Mean tokens}} & \multicolumn{3}{c}{\textbf{Energy}} & \multicolumn{2}{c}{\textbf{Token}} & \multicolumn{2}{c}{\textbf{Solo}} \\
\cmidrule(lr){3-4} \cmidrule(lr){5-7} \cmidrule(lr){8-9} \cmidrule(lr){10-11}
 & & \textbf{Prefill} & \textbf{Decode} & \textbf{$\sum$Solo (J)} & \textbf{Batch (J)} & \textbf{Save} & \textbf{J} & \textbf{L1} & \textbf{J} & \textbf{L1} \\
\midrule
\multicolumn{11}{c}{\textbf{Static Batching}} \\
\midrule
\multirow{4}{*}{Qwen2.5-7B} & Chat & 72 & 421 & 9770 & 1571 & 84\% & 170 & 0.108 & 62 & 0.040 \\
 & LongBench & 921 & 253 & 6083 & 1714 & 72\% & 958 & 0.546 & 259 & 0.150 \\
 & Reasoning & 103 & 273 & 6275 & 1160 & 82\% & 324 & 0.268 & 287 & 0.239 \\
 & Synthetic & 1722 & 198 & 5041 & 1818 & 64\% & 1664 & 0.917 & 246 & 0.135 \\
\midrule
\multirow{4}{*}{Qwen2.5-14B} & Chat & 72 & 399 & 17764 & 2949 & 83\% & 297 & 0.101 & 160 & 0.055 \\
 & LongBench & 921 & 243 & 11505 & 3405 & 70\% & 2110 & 0.607 & 723 & 0.214 \\
 & Reasoning & 103 & 299 & 13457 & 2465 & 82\% & 680 & 0.264 & 546 & 0.215 \\
 & Synthetic & 1722 & 189 & 9175 & 3561 & 61\% & 3287 & 0.925 & 446 & 0.125 \\
\midrule
\multirow{4}{*}{Llama-3.1-8B} & Chat & 79 & 435 & 10715 & 1664 & 84\% & 151 & 0.091 & 41 & 0.025 \\
 & LongBench & 1005 & 267 & 6813 & 1929 & 72\% & 1085 & 0.555 & 289 & 0.143 \\
 & Reasoning & 109 & 239 & 5784 & 1168 & 80\% & 351 & 0.286 & 315 & 0.254 \\
 & Synthetic & 1053 & 233 & 6016 & 1843 & 69\% & 1502 & 0.815 & 287 & 0.156 \\
\midrule
\multirow{4}{*}{Mistral-7B} & Chat & 54 & 395 & 9137 & 1525 & 83\% & 246 & 0.162 & 160 & 0.105 \\
 & LongBench & 1341 & 228 & 5453 & 1800 & 67\% & 1116 & 0.631 & 546 & 0.297 \\
 & Reasoning & 84 & 265 & 5692 & 1280 & 78\% & 368 & 0.282 & 391 & 0.291 \\
 & Synthetic & 1702 & 316 & 7830 & 1967 & 75\% & 935 & 0.476 & 321 & 0.163 \\
\midrule
Mean & All & 691 & 291 & 8532 & 1989 & 77\% & 953 & 0.440 & 317 & 0.163 \\
\midrule
\multicolumn{11}{c}{\textbf{Continuous Batching}} \\
\midrule
\multirow{4}{*}{Qwen2.5-7B} & Chat & 72 & 419 & 9775 & 2309 & 76\% & 342 & 0.149 & 417 & 0.181 \\
 & LongBench & 921 & 253 & 6211 & 2519 & 59\% & 1545 & 0.605 & 442 & 0.175 \\
 & Reasoning & 103 & 275 & 6516 & 1950 & 70\% & 560 & 0.285 & 552 & 0.283 \\
 & Synthetic & 1722 & 199 & 5131 & 2381 & 54\% & 2092 & 0.892 & 390 & 0.162 \\
\midrule
\multirow{4}{*}{Qwen2.5-14B} & Chat & 72 & 402 & 16990 & 3507 & 79\% & 429 & 0.123 & 521 & 0.149 \\
 & LongBench & 921 & 246 & 10883 & 3809 & 65\% & 2430 & 0.622 & 614 & 0.161 \\
 & Reasoning & 103 & 301 & 12707 & 3030 & 76\% & 800 & 0.261 & 766 & 0.251 \\
 & Synthetic & 1722 & 192 & 8887 & 3951 & 56\% & 3591 & 0.914 & 523 & 0.132 \\
\midrule
\multirow{4}{*}{Llama-3.1-8B} & Chat & 79 & 438 & 10319 & 2289 & 78\% & 316 & 0.138 & 381 & 0.166 \\
 & LongBench & 1005 & 276 & 6681 & 2481 & 63\% & 1410 & 0.558 & 452 & 0.182 \\
 & Reasoning & 109 & 244 & 5581 & 1830 & 67\% & 644 & 0.352 & 638 & 0.349 \\
 & Synthetic & 1053 & 235 & 5864 & 2315 & 61\% & 1831 & 0.797 & 488 & 0.211 \\
\midrule
\multirow{4}{*}{Mistral-7B} & Chat & 54 & 391 & 8592 & 2011 & 77\% & 449 & 0.223 & 477 & 0.236 \\
 & LongBench & 1341 & 231 & 5313 & 2238 & 58\% & 1413 & 0.640 & 520 & 0.220 \\
 & Reasoning & 84 & 255 & 5572 & 1643 & 71\% & 420 & 0.252 & 467 & 0.282 \\
 & Synthetic & 1702 & 324 & 7563 & 2453 & 68\% & 1249 & 0.510 & 411 & 0.167 \\
\midrule
Mean & All & 691 & 293 & 8286 & 2545 & 69\% & 1220 & 0.458 & 504 & 0.207 \\
\bottomrule
\end{tabular}
\caption{Full energy breakdown under static and continuous batching, per model and workload. $\sum$Solo is the summed standalone energy, Batch is the measured batch energy, Save is the resulting saving, and Token and Solo give each rule's error in Joules and normalized L1.}
\label{tab:main-full}
\end{table*}

Table~\ref{tab:main-full} gives the full energy breakdown behind the attribution errors in Table~2 of the main paper, per model and workload in both batching regimes. Batching saves 77\% of energy under static batching and 69\% under continuous batching, yet that shared energy is exactly what token attribution misallocates. Summed over requests, token charges deviate from exact Shapley by 953 J on the 1989 J static batch and 1220 J on the 2545 J continuous batch. Since each misallocated joule appears twice in this sum, once as an overcharge and once as an undercharge, roughly a quarter of each batch's energy is charged to the wrong requests.

\subsection{Statistical Significance}
\label{app:significance}

We test that the attribution gaps in \S5.2 of the main paper exceed sampling variation across the 80 groups per regime, using bootstrap 95\% confidence intervals and a paired sign-flip permutation test with $20{,}000$ resamples on per-group L1. Under static batching, mean Token L1 is 0.440 with 95\% CI $[0.374, 0.508]$ and mean Solo L1 is 0.163 with CI $[0.141, 0.186]$. Under continuous batching the corresponding intervals are $[0.395, 0.523]$ and $[0.190, 0.224]$. The Token and Solo intervals are well separated in both regimes, and the paired Token$-$Solo differences, 0.277 static and 0.251 CB, are significant at $p<10^{-4}$.

\Calib's improvement over both baselines (\S5.3 of the main paper) is significant as well. Under static batching, its mean L1 has 95\% CI $[0.102, 0.132]$, the mean differences are $-0.323$ versus Token and $-0.047$ versus Solo, and the paired-permutation test gives $p<10^{-4}$ for both. Under continuous batching, the corresponding CI is $[0.159, 0.197]$ with differences $-0.280$ versus Token and $-0.029$ versus Solo, significant at $p<10^{-4}$ and $p<10^{-3}$ respectively.

\subsection{Case Study: Token Attribution vs.\ Exact Shapley on a Heterogeneous Group}
\label{app:case-study}

\begin{table}[t]
\centering
\setlength{\tabcolsep}{5pt}
\small
\begin{tabular}{lrrrrr}
\toprule
\multirow{2}{*}{\textbf{Request}} & \multicolumn{2}{c}{\textbf{Tokens}} & \multicolumn{3}{c}{\textbf{Energy (J)}} \\
\cmidrule(lr){2-3} \cmidrule(lr){4-6}
 & \textbf{Prefill} & \textbf{Decode} & \textbf{Token} & \textbf{Solo} & \textbf{Shapley} \\
\midrule
lb\_000033 & 8254 & 377 & 2527.6 & 706.3 & \textbf{974.6} \\
lb\_000038 & 1017 & 512 & 447.8 & 868.6 & \textbf{932.3} \\
lb\_000019 & 442 & 512 & 279.4 & 854.5 & \textbf{862.2} \\
lb\_000002 & 281 & 99 & 111.3 & 174.1 & \textbf{108.1} \\
lb\_000046 & 131 & 89 & 64.4 & 154.0 & \textbf{98.0} \\
lb\_000073 & 44 & 50 & 27.5 & 117.0 & \textbf{62.5} \\
lb\_000017 & 39 & 376 & 121.5 & 661.7 & \textbf{533.2} \\
lb\_000075 & 36 & 38 & 21.7 & 65.0 & \textbf{30.3} \\
\midrule
Total & & & 3601.2 & 3601.2 & \textbf{3601.2} \\
\bottomrule
\end{tabular}
\caption{One Qwen2.5-14B LongBench group under static batching. Shapley is the exact reference, in bold, and IDs abbreviate \texttt{longbench\_}.}
\label{tab:case-study}
\end{table}

\begin{table}[t]
\centering
\setlength{\tabcolsep}{5pt}
\small
\begin{tabular}{lrrrrr}
\toprule
\multirow{2}{*}{\textbf{Request}} & \multicolumn{2}{c}{\textbf{Tokens}} & \multicolumn{3}{c}{\textbf{Energy (J)}} \\
\cmidrule(lr){2-3} \cmidrule(lr){4-6}
 & \textbf{Prefill} & \textbf{Decode} & \textbf{Token} & \textbf{Solo} & \textbf{Shapley} \\
\midrule
lb\_000033 & 8254 & 335 & 2858.1 & 831.6 & \textbf{980.1} \\
lb\_000038 & 1017 & 512 & 508.8 & 991.7 & \textbf{1052.6} \\
lb\_000019 & 442 & 512 & 317.5 & 964.5 & \textbf{882.2} \\
lb\_000002 & 281 & 99 & 126.4 & 200.5 & \textbf{158.8} \\
lb\_000046 & 131 & 90 & 73.5 & 178.8 & \textbf{146.6} \\
lb\_000073 & 44 & 67 & 36.9 & 128.9 & \textbf{178.9} \\
lb\_000017 & 39 & 389 & 142.4 & 734.5 & \textbf{542.7} \\
lb\_000075 & 36 & 38 & 24.6 & 57.9 & \textbf{146.4} \\
\midrule
Total & & & 4088.3 & 4088.3 & \textbf{4088.3} \\
\bottomrule
\end{tabular}
\caption{The same group as Table~\ref{tab:case-study}, served under continuous batching with interleaved arrivals. Shapley is the exact reference, in bold.}
\label{tab:case-study-cb}
\end{table}

Tables~\ref{tab:case-study} and~\ref{tab:case-study-cb} show the group with the largest token error across the static runs, served under both regimes. It is the extreme right tail of the per-request error distribution rather than a typical case, with an overall mean per-run maximum token error of 0.207. In both regimes the 8254-prefill request is charged far above its Shapley value (2527.6 vs.\ 974.6 J static, 2858.1 vs.\ 980.1 J under continuous batching), while requests with a few hundred prefill tokens but 512 decode tokens are undercharged. The staggered arrivals raise the batch energy (4088.3 vs.\ 3601.2 J), but the Shapley value of the dominant request is nearly identical across regimes, so the misattribution follows from shared batched execution rather than from the static replay protocol.

\subsection{Calibration Generalization}
\label{app:calib-gen}

Table~\ref{tab:calib-gen} breaks the deployed \Calib estimator down by evaluation split. The within-run split gives the headline number reported in \S5.3 of the main paper (0.116 static). Pooled leave-one-group-out, leave-model-out, and leave-workload-out are progressively harder transfer settings, and across every split and both regimes the estimator stays far below the Token rule. Collapsing the separate prefill and decode channels into a single total feature inflates leave-model-out error from 0.127 to 0.395 (static) and 0.190 to 0.383 (CB), confirming that the asymmetric per-token cost of prefill versus decode, not model capacity, is what transfers across domains. Finally, the estimator transfers across the scheduling regime itself. Fit on static runs and tested on CB it reaches 0.216, and the reverse reaches 0.162, so a calibrator trained offline on static batches remains usable under a live scheduler.

\begin{table}[t]
\centering
\setlength{\tabcolsep}{5pt}
\small
\begin{tabular}{lrr}
\toprule
\textbf{Attribution} & \textbf{Static} & \textbf{CB} \\
\midrule
Token & 0.440 & 0.458 \\
Solo & 0.163 & 0.207 \\
\midrule
\Calib & & \\
\quad within-run & \textbf{0.116} & \textbf{0.177} \\
\quad pooled leave-one-group-out & 0.126 & 0.189 \\
\quad leave-model-out & 0.127 & 0.190 \\
\quad leave-workload-out & 0.185 & 0.216 \\
\bottomrule
\end{tabular}
\caption{Normalized L1 against exact Shapley for \Calib under four evaluation splits, with Token and Solo for reference. Bold marks the lowest L1 per regime.}
\label{tab:calib-gen}
\end{table}

\subsection{Supervision Ablation: Batch-Energy Fit}
\label{app:energy-fit}

Table~\ref{tab:energy-fit} compares supervision signals for the same attribution form, all under the within-run leave-one-group-out protocol of \S5.3 in the main paper. The batch-energy rows fit $\Energy(\Requests) \approx a\sum_i p_i + b\sum_i d_i$ by least squares over the training groups' full-batch energies, one equation per group and no subset replay, and attribute in proportion to $a\,p_i + b\,d_i$ with the same clip-and-renormalize step as \Calib. The overhead variant adds an intercept that is split evenly at attribution time. It performs worse because the intercept is poorly identified from so few equations and absorbs a large share of the batch energy, and splitting that share evenly drags the allocation toward a uniform split, which is far from exact Shapley for heterogeneous groups. The Shapley rows fit the same two token features, or the full feature set, against measured Shapley shares. Batch energy is nonlinear in token totals and a within-run fit sees only four batch equations, which under-identifies the prefill/decode marginal ratio. Pooling more batches helps only modestly, with leave-model-out at 0.259 static and 0.275 CB. Measured Shapley shares expose the marginal structure directly and recover most of the accuracy even with two features.

\begin{table}[t]
\centering
\setlength{\tabcolsep}{5pt}
\small
\begin{tabular}{lrr}
\toprule
\textbf{Supervision signal} & \textbf{Static} & \textbf{CB} \\
\midrule
Batch energies & & \\
\quad proportional $a\,p_i + b\,d_i$ & 0.367 & 0.335 \\
\quad with overhead intercept & 0.536 & 0.431 \\
\midrule
Shapley shares & & \\
\quad same two features & 0.129 & 0.199 \\
\quad full \Calib features & \textbf{0.116} & \textbf{0.177} \\
\bottomrule
\end{tabular}
\caption{Within-run normalized L1 for the same proportional attribution form under different supervision signals. Bold marks the lowest L1 per regime.}
\label{tab:energy-fit}
\end{table}

\subsection{Cross-Hardware Robustness}
\label{app:crosshw}

Table~\ref{tab:crosshw} reports per-workload Token L1 for the same Qwen2.5-7B groups measured on the A800, A40, and H100 under both batching regimes. The per-workload pattern matches the main campaign. Chat is mildest, the long-context and synthetic groups dominate the error, and the ordering is stable across all three GPUs.

\begin{table}[t]
\centering
\setlength{\tabcolsep}{4.5pt}
\small
\begin{tabular}{lrrrrrr}
\toprule
\multirow{2}{*}{\textbf{Workload}} & \multicolumn{3}{c}{\textbf{Static Batching}} & \multicolumn{3}{c}{\textbf{Continuous Batching}} \\
\cmidrule(lr){2-4} \cmidrule(lr){5-7}
 & \textbf{A800} & \textbf{A40} & \textbf{H100} & \textbf{A800} & \textbf{A40} & \textbf{H100} \\
\midrule
Chat & 0.094 & 0.106 & 0.109 & 0.154 & 0.134 & 0.195 \\
LongBench & 0.337 & 0.333 & 0.337 & 0.379 & 0.297 & 0.405 \\
Reasoning & 0.329 & 0.342 & 0.357 & 0.321 & 0.304 & 0.321 \\
Synthetic & 1.107 & 1.119 & 1.200 & 1.110 & 1.097 & 1.113 \\
\midrule
Mean & 0.467 & 0.475 & 0.501 & 0.491 & 0.458 & 0.508 \\
\bottomrule
\end{tabular}
\caption{Token L1 against exact Shapley for the same two Qwen2.5-7B groups per workload on three GPUs, under static and continuous batching.}
\label{tab:crosshw}
\end{table}

\subsection{\boldmath Sampled Shapley at $n{=}16$: Per-Workload Results}
\label{app:sampled-n16}

Table~\ref{tab:sampled-n16-full} breaks Table~3 of the main paper down by workload. The pattern matches the exact $n{=}8$ campaign, with chat mildest and synthetic driving the largest Token errors. On CB chat at $n{=}16$, Token attribution edges out \Calib (0.269 vs.\ 0.282), the one cell where token counts remain the best deployable rule. In every other cell \Calib is the most accurate deployable rule.

\begin{table}[t]
\centering
\setlength{\tabcolsep}{3.5pt}
\small
\begin{tabular}{lrrrrrr}
\toprule
\multirow{2}{*}{\textbf{Workload}} & \multicolumn{3}{c}{\textbf{Static Batching}} & \multicolumn{3}{c}{\textbf{Continuous Batching}} \\
\cmidrule(lr){2-4} \cmidrule(lr){5-7}
 & \textbf{Token} & \textbf{Solo} & \textbf{\Calib} & \textbf{Token} & \textbf{Solo} & \textbf{\Calib} \\
\midrule
\multicolumn{7}{l}{\boldmath\textbf{Exact Shapley reference, $n{=}8$}} \\
\cmidrule(lr){1-7}
Chat & 0.109 & \textbf{0.025} & 0.052 & 0.195 & 0.224 & \textbf{0.194} \\
LongBench & 0.337 & 0.103 & \textbf{0.074} & 0.405 & 0.180 & \textbf{0.122} \\
Reasoning & 0.357 & 0.282 & \textbf{0.227} & 0.321 & 0.315 & \textbf{0.307} \\
Synthetic & 1.200 & \textbf{0.180} & 0.208 & 1.113 & 0.186 & \textbf{0.159} \\
\cmidrule(lr){1-7}
Mean & 0.501 & 0.148 & \textbf{0.140} & 0.508 & 0.226 & \textbf{0.196} \\
\midrule
\multicolumn{7}{l}{\boldmath\textbf{Sampled Shapley reference, $n{=}16$}} \\
\cmidrule(lr){1-7}
Chat & 0.223 & \textbf{0.142} & 0.159 & \textbf{0.269} & 0.319 & 0.282 \\
LongBench & 0.437 & 0.150 & \textbf{0.126} & 0.623 & 0.325 & \textbf{0.282} \\
Reasoning & 0.423 & 0.352 & \textbf{0.298} & 0.215 & 0.233 & \textbf{0.211} \\
Synthetic & 0.819 & 0.227 & \textbf{0.205} & 0.944 & 0.422 & \textbf{0.358} \\
\cmidrule(lr){1-7}
Mean & 0.475 & 0.218 & \textbf{0.197} & 0.512 & 0.325 & \textbf{0.283} \\
\bottomrule
\end{tabular}
\caption{Per-workload breakdown of Table~3 in the main paper (Qwen2.5-7B, H100). Bold marks the lowest measured L1 per workload, regime, and block.}
\label{tab:sampled-n16-full}
\end{table}

\subsection{Model-Scale Robustness}
\label{app:small-model}

Table~\ref{tab:small-model} reports the three attribution rules on Qwen2.5-1.5B under both regimes. Token L1 is large for the long-context and synthetic workloads (0.521 and 0.778 under static batching) and larger on chat than for the 7B model (0.249 vs.\ 0.108). \Calib again recovers most of the gap. Continuous batching gives the same picture. Token L1 stays large (mean 0.454), \Calib is the most accurate rule on every workload (mean 0.204), and Solo sits in between.

\begin{table}[t]
\centering
\setlength{\tabcolsep}{3.5pt}
\small
\begin{tabular}{lrrrrrr}
\toprule
\multirow{2}{*}{\textbf{Workload}} & \multicolumn{3}{c}{\textbf{Static Batching}} & \multicolumn{3}{c}{\textbf{Continuous Batching}} \\
\cmidrule(lr){2-4} \cmidrule(lr){5-7}
 & \textbf{Token} & \textbf{Solo} & \textbf{\Calib} & \textbf{Token} & \textbf{Solo} & \textbf{\Calib} \\
\midrule
Chat & 0.249 & \textbf{0.166} & 0.203 & 0.219 & 0.261 & \textbf{0.205} \\
LongBench & 0.521 & 0.188 & \textbf{0.131} & 0.586 & 0.186 & \textbf{0.156} \\
Reasoning & 0.267 & 0.241 & \textbf{0.180} & 0.269 & 0.281 & \textbf{0.217} \\
Synthetic & 0.778 & 0.156 & \textbf{0.149} & 0.740 & 0.252 & \textbf{0.236} \\
\midrule
Mean & 0.454 & 0.188 & \textbf{0.166} & 0.454 & 0.245 & \textbf{0.204} \\
\bottomrule
\end{tabular}
\caption{Normalized L1 against exact Shapley for the three rules on Qwen2.5-1.5B, under static and continuous batching. Bold marks the lowest L1 per workload and regime.}
\label{tab:small-model}
\end{table}

\subsection{Reproducibility Notes}

The full campaign used vLLM 0.8.2, PyTorch 2.6.0, and Transformers 4.51.3 on Linux clusters under SLURM, completing all 16 model/workload runs in both batching regimes (80 groups per regime, 40,800 subset measurements). The sampled-Shapley campaign adds 83,456 prefix-coalition replays at $n{=}16$ on the H100. Group construction and permutation schedules use fixed seeds and are saved alongside the measurements. Raw power traces are saved for every static subset, and the CB runs persist integrated per-subset energies. Per-group attribution outputs are saved for every subset in both regimes. The numbers in the paper are regenerated from the raw measurements by the analysis scripts, which compute the bootstrap confidence intervals, paired permutation tests, group-size sweep, repeatability, and calibration ablations, and verify the reported values against the raw data.

\end{document}